\documentclass[12pt]{article}
\usepackage[utf8]{inputenc}
\usepackage[T1]{fontenc}
\usepackage{a4wide}

\usepackage{hyperref}
\usepackage{graphicx}
\usepackage[usenames]{color}
\usepackage{amsmath,amsfonts,amssymb}
\usepackage{mathtext,enumerate,float}
\usepackage[english]{}
\usepackage[all]{xy}
\usepackage{subfig}
\usepackage{url}

\DeclareMathOperator*{\argmax}{arg\,max}

\DeclareMathOperator*{\idx}{idx}

\DeclareMathOperator*{\AUCH}{AUCH}
\DeclareMathOperator*{\tf}{tf}

\DeclareMathOperator*{\idf}{idf}

\DeclareMathOperator*{\const}{const}

\DeclareMathOperator*{\dBern}{Bern}

\DeclareMathOperator*{\dKL}{KL}
\DeclareMathOperator*{\diag}{diag}

\newcommand{\dd}[1]{\mathrm{d}{#1}}

\newcommand{\precision}{p@k}
\newcommand{\dcg}{\mathrm{DCG}}
\newcommand{\rel}{\mathrm{rel}}

\newcommand{\mmatr}{{\mathbf{M}}}

\newcommand{\imatr}{{\mathbf{I}}}
\newcommand{\vmatr}{{\mathbf{V}}}
\newcommand{\wmatr}{{\mathbf{W}}}

\newcommand{\zmatr}{{\mathbf{Z}}}
\newcommand{\zmatrt}{{\tilde{\mathbf{Z}}}}

\newcommand{\lambdamatr}{{\mathbf{\Lambda}}}

\newcommand{\hb}{{\mathbf{h}}}
\newcommand{\mb}{{\mathbf{m}}}
\newcommand{\pb}{{\mathbf{p}}}

\newcommand{\xb}{{\mathbf{x}}}
\newcommand{\xt}{{\tilde{x}}}
\newcommand{\xbt}{\tilde{{\mathbf{x}}}}
\newcommand{\yb}{{\mathbf{y}}}

\newcommand{\zt}{{\tilde{z}}}

\newcommand{\mub}{{\boldsymbol{\mu}}}
\newcommand{\alphab}{{\boldsymbol{\alpha}}}
\newcommand{\thetab}{{\boldsymbol{\theta}}}
\newcommand{\iotab}{\boldsymbol{\iota}}

\newcommand{\xib}{\boldsymbol{\xi}}
\newcommand{\xibt}{\tilde{\boldsymbol{\xi}}}
\newcommand{\xit}{\tilde{\xi}}

\newcommand{\shi}{s_{\mathsf{h}}}

\newcommand{\entr}{\mathsf{H}}

\newcommand{\prob}{p}
\newcommand{\expec}{\mathsf{E}}

\newcommand{\Lmc}{{\mathcal{L}}}

\newcommand{\Nmc}{{\mathcal{N}}}

\newcommand{\Tmc}{{\mathcal{T}}}
\newcommand{\Vmc}{{\mathcal{V}}}
\newcommand{\Wmc}{{\mathcal{W}}}

\newcommand{\T}{^{\text{\tiny\sffamily\upshape\mdseries T}}}
\newcommand{\deist}{\mathbb{R}}

\begin{document}
\title{Hierarchical thematic classification of major conference proceedings}
\author{Arsentii A. Kuzmin, Alexander A. Aduenko, and Vadim V. Strijov\footnote{E-mail: vadim@m1p.org}}
\maketitle

\begin{abstract}
In this paper we develop a decision support system for the hierarchical text classification. We consider text collections with fixed hierarchical structure of topics given by experts in the form of a tree. The system sorts the topics by relevance to a given document. The experts choose one of the most relevant topic to finish the classification. We propose a weighted hierarchical similarity function to calculate topic relevance. The function calculates similarity of a document and a tree branch. The weights in this function determine word importance. We use the entropy of words to estimate the weights. 

The proposed hierarchical similarity function formulate a joint hierarchical thematic classification probability model of the document topics, parameters, and hyperparameters. The variational bayesian inference gives a closed form EM algorithm. The EM algorithm estimates the parameters and calculates the probability of a topic for a given document. Compared to hierarchical multiclass SVM, hierarchical PLSA with adaptive regularization, and hierarchical naive bayes, the weighted hierarchical similarity function has better improvement in ranking accuracy in an abstracts collection of a major conference EURO and a web sites collection of industrial companies.
\end{abstract}

\section{Introduction}
A thematic model of a text collection is a map, which determines a set of topics from a given hierarchical structure of topics for each document from the collection. The text collections are scientific abstracts~\cite{Joachims1998}, conference procedeings, text messages from social networks~\cite{SchedlMadonnaSimilarity2012}, web sites~\cite{Xue:2008:DCL:1390334.1390440}, patents description, and news articles~\cite{ikonomakis2005text,feature_selection}. The thematic model helps to search through collections efficiently. But the model constructing is often labor-intensive. Some collections already have a structure and a subset of documents that partly-classified by experts. To simplify the procedure of expert classification the authors propose an algorithm that ranks collection's topics for a given document. On user demand the algorithm puts a new document into the topic with the highest rank.

This paper investigates the thematic modeling problem for partialy-labled collectoions with fixed expert tree structure of topics~\cite{DBLP:conf/icml/McCallumRMN98,DBLP:conf/iconip/KuznetsovCAGS15,hierarchicalSVM2007}. In the tree structure, the leaf topic of a document determines the topics for this document on the other levels of hierarchy. So a required solution is a map, which determines ranks of the leaf topics for a given document. The ranks of the expert topics determine the quality of the solution. 

A relevance of a leaf topic to a given document determines the rank of the topic. The relevance is determined by the value of a discriminant function or the probability estimate of a discriminative or generative model. For a non-hierarchical classification problem~\cite{Joachims:1997:PAR:645526.657278,Joachims1998,Li2006} the SVM, kNN, and Neural Networks return a value of the discriminant function. In~\cite{Frank:2006:NBT:2089856.2089908,Genkin:August2007:0040-1706:291,dpm2010} the Naive Byaes, Multinomial logistic regression, and dPM models give probability estimation for a topic of a given document.

In the hierarchical classification the ``top--down'' aproach~\cite{hierarchicalSVM2007} gives results better than the non-hierarchical classification among leaf-level topics. Starting from the top of the hierarchy it relates a document to one of the children topics using non-hierarchical classification methods~\cite{hierarchy_divisime,Xue:2008:DCL:1390334.1390440}. It's drawback is that a misclassification at the top level immediately leads to a misclassification on the leaf level. To rank the leaf topics using ``top-down'' approach the algorithm sorts them according to the ranks of the parent topics on each level. An alternative way is to consider all clusters of the tree branch at once~\cite{DBLP:conf/icml/McCallumRMN98,Tsochantaridis:2005:LMM:1046920.1088722}.

Taking into account the word importance improves the classification quality. Common approach is to change a frequency-based document description to~$\tf\cdot\idf$ features~\cite{Salton:1988:TAA:54259.54260}. Another approach is to optimize a weighted metric or similarity function~\cite{LeischKCentroids2006,minkowskiMetricWeigthted2012,YihWeightedSimilarity2009,6083923}. 
The disadvantage of the last approach is a huge number of optimization parameters that equals the dictionary size of collection. In~\cite{Largeron:2011:EBF:1982185.1982389} authors use the entropy to estimate the importance of words and reduce the number of parameters. In this paper we improve this approach and generalize it to hierarchical case.

We propose a weighted hierarchical similarity function of a document and a branch of a cluster tree. The function considers the  word importance and hierarchical structure of the collection. We put priors on parameters of the similarity function to regularize them and take into account our assumptions about their default values. The similarity function formulate a joint hierarchical thematic classification probability model of document topics, parameters, and hyperparameters. We use the variational bayesian inference~\cite{DBLP:journals/corr/BleiKM16,DBLP:conf/icml/GershmanHB12,bishop2006pattern} to derive EM algorithm and estimate probability of topics for unlabeled documents.

We consider a process of constructing thematic model of major conference ``European Conference on Operational Research~(EURO)'' as the example of thematic modeling task. A program committee builds the thematic model for this conference from a set of received abstracts every year. Structure of this model consists of~$26$ major Areas, each Area consists of~$10-15$ Streams, each stream consists of~$5-10$ Sessions, and each Session consists of four talks~\ref{fig:euro_structure}. Participants send short abstracts to program committee to apply. There are two types of the participants: invited and new ones. The invited participants have already determined session, so the collection of abstracts is partly-labeled. For each new one participant, the program committee should choose the most relevant session according to his abstract and the conference structure. The program committee invites up to~$200$ experts from  different research areas to construct the thematic model.

We construct decision support system for creating thematic model of the conference, which gives the expert a ranked list of possible clusters for a given document. We use expert models of this conference from the previous years to estimate the parameters and compare the quality of proposed algorithm with commonly used classification methods.  

\begin{minipage}[t]{0.46\linewidth}
\centering
\begin{figure}[H]
\includegraphics[width=\textwidth]{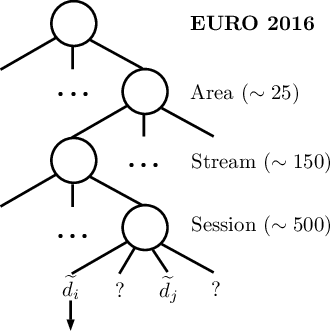}
\caption{Structure of EURO conference.}
\label{fig:euro_structure}
\end{figure}
\end{minipage}
\hfill
\begin{minipage}[t]{0.35\linewidth}
\centering
\begin{figure}[H]
\includegraphics[width=\textwidth]{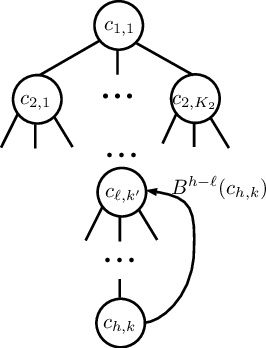}
\caption{Basic notation in hierarchical structure of the collection.}
\label{fig:cluster_hierarchy_1}
\end{figure}
\end{minipage}

\section{Weighted hierarchical similarity function}
Let~$w$ be a word. A document~$d$ is an unordered set of words~$\{w_1, w_2, \dots, w_{|d|}\}$. A document collection~$D$ is an unordered set of documents
\begin{equation}
D = \{d_1, d_2, \dots, d_{|D|}\}.
\end{equation} 
A dictionary~$W$ of the collection~$D$ is an ordered set of unique words~$w$ form the collection~$D$. Each document~$d_n$ is represented by a real value vector~$\xb_n$. The element~$x_{i,n}$ of~$\xb_n$ equals to the number of words~$w_i$ in the document~$d_n$. 

A cluster~$c$ is a subset of documents from the collection~$D$. The experts define a collection structure as a graph of topics. In this paper we consider only trees as the possible collection structures. Each node (leaf) of the topic tree corresponds to a cluster~$c$ of documents from this topic. Let~$h$ be a height of the tree. Indexes~$\ell$ and~$k$ of a cluster~$c_{\ell,k}$ denote level in the tree~$\ell$ and index on this level~$k$. Cluster~$c_{1,1}$ is a root of the tree. Let~$K_{\ell}$ be number of clusters on the level~$\ell$. Cluster~$c_1$ is a parent cluster for~$c_2$ if it contains all documents~$d$ from~$c_2$. Then cluster~$c_2$ is a child cluster of~$c_1$. Let~$B$ be an operator that returns parent cluster of a given cluster. We use~$B$ operator~$h-\ell$ times~$B^{h-\ell}(c_{h,k})$ to get parent cluster on the level~$\ell$ of the lowest level cluster~$c_{h,k}$~(see. fig.~\ref{fig:cluster_hierarchy_1}).

Let~$c(d)$ be an expert cluster of document~$d$ on the lowest level~$h$. Matrix~$\zmatr$ determines the expert clusters on the lowest level for all documents:
\begin{equation}
\label{eq:zmatrix_def}
z_{nk} = [d_n \in c_{h,k}], \quad z_{nk}\text{~~is the element of~} \zmatr.
\end{equation}
Leaf cluster determines classification of the document on all other levels of the hierarchy. So matrix~$\zmatr$ determines a whole expert thematic model.

\vspace{0.3cm}
\textbf{Quality criterion for hierarchical ranking.} For each document the algorithm ranks all leaf clusters according to their relevance. Then the expert choose one best cluster from the ranked set. The rank of the chosen cluster determines quality of the ranking: the lower the rank, the better the quality.

Let~$S^{K_h}$ be a permutation set of order~$K_h$. Let~$R$ be a relevance operator. Relevance operator maps each document $\xb \in \deist^{|W|}$ to clusters permutation~$q(\xb) \in S^{K_h}$ of the level~$h$. The clusters in permutation~$q(\xb)$ are sorted by relevance to document~$\xb$ in descending order. The rank of each cluster equals to position in permutation. The goal is to find such operator~$R(\xb)$, that has the best quality on the expert classification~$\zmatr$:
$$
\AUCH(R, D, \zmatr) \rightarrow \max,
$$
where~$\AUCH(R, D, \zmatr)$ is a quality function, which depends on  ranks of the expert clusters.

\begin{figure}[H]
\centering
\includegraphics[width=0.35\textwidth]{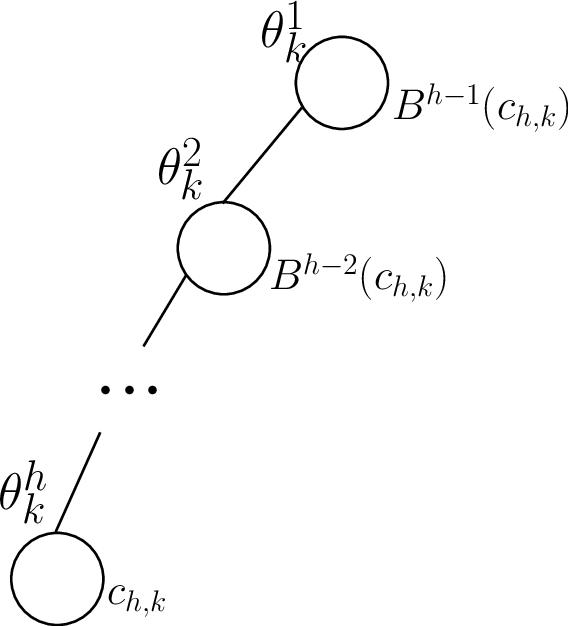}
\caption{A branch with the number~$k$ of a cluster hierarchy. The value of~$\theta_k^\ell$ denote the weight of the cluster~$c_{\ell, k}$ in the branch.}
\label{fig:cluster_hierarchy}
\end{figure}

A common ranking quality criterion is the discounted cumulative gain~$\dcg_k$ and~$\precision$:
\begin{equation}
\mathrm{DCG}_k = \rel_1 + \sum_{i = 1}^k \frac{\rel_i}{\log_2 i}, \quad \precision = \frac{r_k}{k},
\end{equation}  
where~$\rel_i$ is a relevance of the cluster with rank~$i$, and~$r_i$ is the number of relevant clusters among the first~$i$ clusters in the permutation~$q(\xb)$. In our case, an expert always selects single cluster. If he selects the cluster with the rank~$j$, then~$\rel_i = 1$ if~$i = j$ and~$0$ otherwise. Same for $r_k$: if the rank of the selected cluster is less then~$k$, then $r_k = 1$ and $0$ otherwise.

$\dcg_k$ and $\precision$ are too detailed in the context of our problem. We propose a simplified quality criterion~AUCH instead. By~$\mathrm{pos}\bigl(R(\xb_n),\,c(\xb_n)\bigr)$ denote the rank of the expert cluster~$c(\xb_n)$ of document~$\xb_n$ according to the permutation~$R(\xb_n)$. We introduce a monotone increasing cumulative histogram of documents with respect to their expert ranks:
\begin{equation}
\#\{n: \; \mathrm{pos}\bigl(R(\xb_n),\,c(\xb_n)\bigr) \leq k\}, \quad k \in [1,\:K_h].
\label{eq:histogram}
\end{equation}
The quality criterion~$\AUCH(R)\in [0,\:1]$~(from Area Under Cumulative Histogram) equals the area under this histogram, normalized by the number of documents and clusters:
\begin{equation}
\AUCH(R) = \dfrac{1}{K_h |D|}\sum_{k=1}^{K_h}{\#\{n: \; \mathrm{pos}\bigl(R(\xb_n),\,c(\xb_n)\bigr) \leq k\}}.
\label{eq:auch}
\end{equation}
Fig.~\ref{fig:algo_comp_area_stream_industry} illustrates an example of envelope curve for the histogram. The value~$\AUCH(R) = 1$ corresponds to the optimal case, when the expert cluster of each document stands on the first position in the permutation.

\paragraph{Weighted similarity of document and cluster.} To rank the clusters one should estimate relevance value of each cluster~$c_{\ell,k}$ to a given document~$\xb$. Let similarity~$s(\xb,\:\yb)$ of documents~$\xb$ and~$\yb$ be a weighted cosin similarity function
\begin{equation}
\label{similar_func_docs}
s(\xb,\:\yb)=\dfrac{\xb^{\T} \lambdamatr \yb}{\sqrt{\xb^{\T}\lambdamatr \xb} \sqrt{\yb^{\T}\lambdamatr \yb}}.
\end{equation}
If the denominator equals zero, the similarity also equals zero. A symmetric non-negative definite matrix~$\lambdamatr$ determines importance of the words. In this paper we consider a diagonal~$\lambdamatr$, because optimization of all its~$|W| \times |W|$ elements leads to inadequate increase of model complexity. We normalize all documents to unclutter notation:
\begin{equation}
\xb \mapsto \dfrac{\xb}{\sqrt{\xb^{\T}\lambdamatr \xb}}, \quad s(\xb,\:\yb) = \xb^{\T} \lambdamatr \yb.
\label{similar_func}
\end{equation}

Let~$\mub(c_{\ell, k})$ be the mean vector of a cluster~$c_{\ell, k}$
\begin{equation}
\mub(c_{\ell, k}) = \frac{1}{|c_{\ell,k}|}\sum_{\xb \in c_{\ell,k}} \xb.
\end{equation}
A similarity~$s(\xb,\:c_{\ell,k})$ of a document and a cluster is the similarity function value~\eqref{similar_func_docs} of the document vector~$\xb$ and the mean vector~$\mub(c_{\ell,k})$ of the cluster:
\begin{equation}
s(\xb,\:c_{\ell,k}) = \xb^{\T} \lambdamatr \mub(c_{\ell,k}).
\label{similar_func}
\end{equation}

\paragraph{Entropy model of word importance.} The number of weight parameters in the diagonal matrix~$\lambdamatr$ equals the size of the dictionary~$|W|$. We propose an entropy model to decrease the number of parameters and avoid overfitting. It maps entropy of a word~$w_m$ to the importance~$\lambda_m$ of this word.

Words that separate clusters are the most important for classification. To understand what does it mean for a word to separate clusters, lets consider the following example. Let all documents from a cluster~$c_{\ell, k}$ contain a word~$w$, and all documents from the other clusters do not contain~$w$. Then it is adequate to classify a new unlabeled document with the word~$w$ to the class~$c_{\ell, k}$. Entropy approach formalizes this idea.

Let~$p_{m,k}^{\ell} = p(c_{\ell, k}| w_m)$ be a probability of cluster~$c_{\ell, k}$ given word~$w_m$. We can estimate~$p_{m,k}^{\ell}$
\begin{equation}
\label{entropy_estimate}
\pb^{\ell}_m = \bigl[\mu(c_{\ell,1})_m,\:\ldots,\:\mu(c_{\ell,K_{\ell}})_m\bigr]\T, \quad \pb^{\ell}_m \mapsto \frac{\pb^{\ell}_m}{\|\pb^{\ell}_m\|_1},
\end{equation}
where~$\mub(c_{\ell, k})_m$ is an~$m$-th component of the cluster's~$c_{\ell, k}$ mean vector. We define entropy of word~$w_m$ according to expert classification on the level~$\ell$ as
\begin{equation}
\label{entrophy}
\entr^{\ell}(w_m) = -\sum_{k = 1}^{K_{\ell}}{p^{\ell}_{m, k}\log(p^{\ell}_{m, k})}.
\end{equation}
The smallest entropy value~$\entr^{\ell}(w_m) = 0$ corresponds to the case, when word~$w_m$ occurs only in documents of one cluster and separate this cluster from the others. Maximum entropy value corresponds to uniform distribution of word~$w_m$ over all clusters,~$p^{\ell}_{m, k} = \mathrm{const}$. In this case~$w_m$ is unimportant word.

In case of hierarchical structure, we calculate entropy of the word according to each level~$\ell$. We combine these values to get the importance value~$\lambda_m$ of the word~$w_m$:
\begin{equation}
\label{lambda_model_old}
\lambda_m = 1 + \sum_{\ell=1}^{h}\alpha_{\ell} \log\bigl(1 + \entr^{\ell}(w_m)\bigr).
\end{equation}
Parameter~$\alpha_{\ell}$ determines influence of words entropy on the level~$\ell$ to importance of the words. Expression~$\log\bigl(1 + \entr^{\ell}(w_m)\bigr)$ doesn't contain any variables so  we calculate it for each word and level~$\ell$ and denote
$$
\iota_{m\ell} \equiv \log\bigl(1 + \entr^{\ell}(w_m)\bigr). 
$$
Then the model~\eqref{lambda_model_old} takes form
\begin{equation}
\label{lambda_model}
\lambda_m = 1 + \alphab\T \iotab_m.
\end{equation}

\paragraph{Hierarchical ranking.} One solution for hierarchical ranking is top-down approach. Let~$C_h(c_{\ell, k})$ be a set of level~$h$ clusters, that are children clusters for~$c_{\ell, k}$. Let~$\idx(c_{\ell, k})$ be the rank of the cluster~$c_{\ell, k}$ on the level~$\ell$. 
We go down from the root of the tree and on each level~$\ell$  rearrange the lowest level clusters~$C_h(c_{\ell, k})$ in the way to preserve the condition
$$
\idx(c_{\ell, k_1}) < \idx(c_{\ell, k_2}) \Rightarrow \idx(c_{h, k_1'}) < \idx(c_{h, k_2'}), \quad \forall k_1, ~k_2, ~c_{h, k_1'} \in C_h(c_{\ell, k_1}),~c_{h, k_2'} \in C_h(c_{\ell, k_2}).
$$
This approach remains top-down disadvantage: wrong ranking on the high level of the tree immediately leads to wrong ranking on the lowest level. Let~$i$ be the rank of the cluster~$c_{2, \hat{k}}$ that contains expert cluster~$c(\xb)$ for a given document. Then the rank of the expert cluster for this document on the lowest level~$h$ will be at least
$$
\idx(c(\xb)) > \sum_{k\;:\;\idx(c_{2, k}) < i} |C_h(c_{2, k})|.
$$

We propose hierarchical similarity function to deal with this problem. It considers similarity with all clusters of the tree branch at once and ranks the set of tree branches instead of single clusters on each level. Tree branches and lowest level clusters have one-to-one correspondence, so we further do not differ ranking of branches and ranking of the lowest level clusters. We also call branch that contains lowest level cluster~$c_{h,k}$ as a branch number~$k$, see. fig.~\ref{fig:cluster_hierarchy}.

Let~$\thetab_k \in \deist^{h}$ be a weight vector for a branch~$k$. Element~$\theta_k^\ell$ of this vector denotes importance of the level~$\ell$ cluster in the branch for classification. In general, if branches~$k_1, \dots, k_n$ contain internal cluster~$c_{\ell,k}$, then there is a set of weights~$\{\theta_{k_1}^{\ell}, \dots, \theta_{k_n}^{\ell} \}$ that corresponds to this cluster and these weights can be different.

Let~$\mub_{\ell,k}$ be the mean vector of the parent cluster~$B^{h-\ell}(c_{h,k})$ of cluster~$c_{h,k}$
$$
\mub_{\ell,k} = \mub\bigl( B^{h-\ell}(c_{h,k})\bigr).
$$
Placing all these vectors for branch~$k$ together gives us the mean vector matrix~$\mmatr_k$. Column~$\ell$ of this matrix corresponds to parent cluster~$B^{h-\ell}(c_{h,k})$ and equals~$\mub_{\ell,k}$:
$$
\mmatr_k = [\mub_{1,k}, \dots, \mub_{h,k}].
$$

We define weighted hierarchical similarity~$\shi(\xb,\: c_{h,\,k})$ of a document~$\xb$ and the lowest level cluster~$c_{h,\,k}$ as a weighted sum of similarities of the document~$\xb$ and clusters~$c_{\ell,\,k}$ of the branch~$k$
\begin{equation}
\label{effective_object_to_cluster_similarity}
\shi(\xb,\: c_{h,\,k}) = \sum_{\ell = 1}^{h} \theta^{\ell}_k s\bigl(\xb,\:B^{h - \ell}(c_{h,\,k})\bigr) \equiv \sum_{\ell = 1}^h \theta^{\ell}_k \xb\T \lambdamatr \mub_{\ell, k} \equiv \xb\T \lambdamatr \mmatr_k \thetab_k.
\end{equation}
The document should be similar to all clusters of the branch to be similar with the lowest level cluster of this branch.


\section{Model and parameters estimation}
\label{sec:greedy_approach}

Hierarchical similarity function contains two sets of parameters: parameter vector~$\alphab$ of the entropy model and set of branches weight vectors~$\thetab = \{\thetab_k\}$. In this section we describe a way to optimize these parameters directly maximizing the quality~$\AUCH$~\eqref{eq:auch} of relevance operator.

Let~$D_{\Vmc_0} \cup D_{\Vmc_1} \cup D_{\Vmc_2}$ be a disjoint subsets of a training set~$D$. We set initial values of parameters
\begin{equation}
\label{eq:parameters_greedy}
\alphab = \mathbf{0}, \quad \thetab_k = \left[\frac{1}{h}, \ldots, \frac{1}{h}\right].
\end{equation}
Optimization algorithm alternates between the following steps: 
\begin{enumerate}
\item[1)] find optimal values of~$\alphab$ given fixed values of~$\thetab_k$ using subset~$D_{\Vmc_1}$,
\item[2)] find optimal values of~$\thetab_k$ given fixed values of~$\alphab$ using subset~$D_{\Vmc_2}$.
\end{enumerate}
In the next subsections we describe each of these steps in more details.

\paragraph{Optimization of entropy model parameters~$\alphab$.} We calculate mean vectors~$\{\mub(c_{\ell, k})\}$ of the clusters using subset~$D_{\Vmc_0}$. These vectors give estimates of words entropy using~\eqref{entropy_estimate} and~\eqref{entrophy} for each level of the hierarchy. We find optimal~$\alpha_1,\,\ldots,\,\alpha_h$ parameters of the entropy model~\eqref{lambda_model} by solving~$\AUCH(R)$~\eqref{eq:auch} maximization task using training subset~$D_{\Vmc_1}$:
\begin{equation}
\label{lambda_optimize}
\alphab^* = \mathop{\arg \max}\limits_{\alphab} {\AUCH(R)}.
\end{equation}
For the small number of levels~$h$ this could be done by grid search.
Changing~$\alphab$ leads to new~$\lambdamatr$ value, so after each iteration we should renormalize document vectors~$\xb$ to preserve~$\xb\T \lambdamatr \xb = 1$ condition and recalculate mean vectors~$\mub(c_{\ell, k})$.

\paragraph{Optimization of the weight vectors~$\{\thetab_k\}$.} Given training subset~$D_{\Vmc_2}$ we need to find set of~$\{\thetab_k\}$ that maximizes hierarchical similarity of documents from~$D_{\Vmc_2}$ with their expert clusters. We keep~$\alphab$ fixed, so values of~~$\lambdamatr$ and~$\xb\T\lambdamatr\mmatr_k$ are known for all documents~$\xb$. This leads to convex quadratic programming task which is solved using interior point method:
\begin{equation}
\thetab_k^* = \argmax_{\thetab_k} \sum_{\xb \in c_{h, k}} \xb\T\lambdamatr \mmatr_k \thetab_k + \psi \|\thetab_k - \hb \|_2^2, 
\label{eq_sim_hierarchy_theta_optimization}
\end{equation}
\begin{equation}
\|\thetab_k\|_1 = 1, \quad \thetab_k \geq \mathbf{0},~~k \in \{1 \ldots K_h\}, \quad \hb = \left[\frac{1}{h}, \dots, \frac{1}{h}\right]\T,
\label{eq_sim_hierarchy_theta_optimization_bounds}
\end{equation}
where~$\psi$~--~is the regularization parameter. We should keep~$\psi \neq 0$, otherwise we face overfitting, because~\eqref{eq_sim_hierarchy_theta_optimization} becomes a linear programming task and optimal solution will be a vertex of the simplex which makes one element of each~$\thetab_k$ equals~$1$ and all other~$0$.

The complexity of this algorithm is~
\begin{equation}
\label{eq:complexity_1}
O(b a^h|D||W|h K_h),
\end{equation} 
where~$b$ is the number of steps~2 and~3, and~$a$ is the number of~$\alpha_{\ell}$ different values in the optimization grid. Experiments showed convergence in~$b \sim 10$ steps.

\section{Bayesian approach in parameters estimation}
\label{sec:bayes}
Quality criterion~$\AUCH$~\eqref{eq:auch} is based on ranking and has discrete values. It restricts the set of possible optimization approaches. Still the valid one from the section~\ref{sec:greedy_approach} has exponential increase of complexity~\eqref{eq:complexity_1} with number of hierarchy levels~$h$. It also involves dividing training set into three subsets and decreases number of objects available for optimization with respect to each set of parameters~$\thetab$ and~$\alphab$. In this section we use likelihood instead of~$\AUCH$ to use more effective optimization methods. We define likelihood of document class matrix~$\zmatr$ as
\begin{equation}
\label{eq:neural_classifier_log_likelihood}
L(\zmatr | \thetab, \alphab) = \prod_{n = 1}^N\prod_{k = 1}^{K_h} \prob^{z_{nk}}(z_{nk} = 1|\xb_n, \thetab_k, \alphab),
\end{equation}
where probability of class~$c_{h, k}$ given document~$\xb_n$ is calculated using softmax function of weighted hierarchical similarity values~$s_{n, k}$ of document~$\xb$ and tree branch~$k$:
\begin{equation}
\label{eq:softmax_prob}
p(z_{nk} = 1| \xb_n, \thetab, \alphab) =  \frac{\exp(s_{n, k})}{\sum_{k' = 1}^{k_h} \exp(s_{n, k'})}, \quad s_{n, k} = s_{\mathsf{h}}(\xb_n, c_{h,k}|\thetab_k, \alphab).
\end{equation}
We assume that parameters~$\{\thetab_k\}$ and~$\alphab$ are random variables with the following distributions
\begin{equation}
\label{eq:bayes_prior_alpha_theta}
p(\alphab) = \Nmc(\alphab | \mathbf{0}, a^{-1}\imatr), \quad p(\thetab_k) = \Nmc(\thetab_k | \mb_k, \vmatr_k^{-1}).
\end{equation}
These priors normalize values of~$\alphab$ and $\{\thetab_k\}$ and take into account our assumptions about them. Vector~$\alphab$ determines influence of words entropy on words importance. Zero value of~$\alphab$ leads to equal importance of all words. Weights vector~$\thetab_k$ determines the weight of clusters in the branch~$k$ and has unknown expectation and covariance matrix. We put another prior on these hyperparameters
\begin{equation}
\label{eq:bayes_prior_m_v}
p(\mb_k| \vmatr_k ) = \Nmc(\mb_k | \mb_0, (b\vmatr_k)^{-1}), \quad p(\vmatr_k) = \Wmc(\vmatr_k | \wmatr, \nu),
\end{equation}
where $\Wmc$ is a Wishart distribution. Mean vector~$\mb_0$ sets initial assumption that clusters of each branch has the same weight~$\mb_{0,k} = 1/h$. Same idea was used for regularization in the previous section~\eqref{eq:parameters_greedy}. The difference from regularization~\eqref{eq_sim_hierarchy_theta_optimization_bounds} is that elements of~$\thetab_k$ now don't have to sum into~$1$. Still we want to preserve assumption that increase of weight of one cluster in the branch leads to decrease of others' weight. Wishart parameter matrix~$\wmatr$ determines covariance matrices~$(b\vmatr_k)^{-1}$ of~$\thetab_k$. We defines~$\wmatr$ initial value in that manner to get negative correlations between~$\thetab_k$ elements. We propose joint model of document classes~$\zmatr$, parameters~$\thetab, \alphab$, and hyperparameters~$\mb, \vmatr$ as
\begin{equation}
\label{eq:bayes_prob_model_full}
p(\zmatr,\thetab, \mb, \vmatr, \alphab ) = L(\zmatr | \thetab, \alphab)p(\thetab| \mb, \vmatr)p(\mb| \vmatr )p(\vmatr)p(\alphab).
\end{equation}

\paragraph{Estimation of cluster probability given document.} Let~$\zmatrt$ be the class matrix for unlabeled documents~\eqref{eq:zmatrix_def}. The relevance operator~$R$ ranks clusters of the lowest level according to probability of the cluster given document. We use model~\eqref{eq:bayes_prob_model_full} to find posterior distribution of hierarchical similarity parameters~$\alphab$ and~$\{\thetab_k\}$ and estimate these probabilities. Two possible types of estimate are:~1)~use maximum posterior values of parameters~$\thetab_k^{\text{MAP}}, \alphab^{\text{MAP}}$ and calculate probability as a softmax value~\eqref{eq:MAP_estimation} of similarities,~2)~calculate evidence estimate~\eqref{eq:evidence_estimation}. In this paper we use the second approach because it takes into account the shape of the posterior distribution and gives better estimates.
\begin{equation}
p(\zt_{tk} = 1|\xbt_t) = p(\zt_{tk}|\xbt_t, \thetab_k^{\text{MAP}}, \alphab^{\text{MAP}})
\label{eq:MAP_estimation}
\end{equation}
\begin{equation}
p(\zt_{tk} = 1|\xbt_t) = \int p(\zt_{tk} |\xbt_t, \thetab, \alphab ) p(\thetab, \alphab| \zmatr) \dd \thetab \dd \alphab
\label{eq:evidence_estimation}
\end{equation}

It is not possible to calculate posterior distribution of parameters due to non-linearity of likelihood~$L$~\eqref{eq:neural_classifier_log_likelihood} on~$\thetab$ and~$\alphab$. We use variational inference to get posterior estimate~\cite{DBLP:journals/corr/BleiKM16,DBLP:conf/icml/GershmanHB12,Hoffman:2013:SVI:2502581.2502622}~$q$. Integration~\eqref{eq:evidence_estimation} also does not have a closed form solution due to softmax function structure. To avoid multiple similar approximations, we at once approximate joint posterior distribution~$p(\zmatrt, \thetab, \mb, \vmatr, \alphab| \zmatr)$ of unlabeled document classes~$\zmatrt$, parameters, and hyperparameters instead of regular posterior~$p(\thetab, \mb, \vmatr, \alphab| \zmatr)$. Joint posterior distribution is the whole expression under the integral~\eqref{eq:evidence_estimation} and it's approximation allows us to calculate integral analytically.

Joint distribution of proposed model~\eqref{eq:bayes_prob_model_full} and unlabeled document classes~$\zmatrt$ is defined by the following equation
\begin{equation}
\label{eq:bayes_prob_model_predictive}
p(\zmatrt, \zmatr, \thetab, \mb, \vmatr, \alphab ) = p(\zmatrt | \thetab, \alphab) p(\zmatr, \thetab, \alphab, \mb, \vmatr).
\end{equation}
Let~$q(\zmatrt, \thetab, \mb, \vmatr, \alphab)$ be an approximation of joint posterior~$p(\zmatrt, \thetab, \mb, \vmatr, \alphab| \zmatr)$. We use mean field approximation~\eqref{eq:bayes_joint_posterior_factorization_q} and search for optimal~$q$ that minimizes $\dKL$ divergence:
\begin{equation}
\dKL(q(\zmatrt, \thetab, \mb, \vmatr, \alphab) \| p(\zmatrt, \thetab, \mb, \vmatr, \alphab| \zmatr)) \rightarrow \min_{q\,=\,q(\thetab)q(\alphab, \mb, \vmatr)q(\zmatrt)}.
\label{eq:kl_div_joint_posterior}
\end{equation}
\begin{equation}
\label{eq:bayes_joint_posterior_factorization_q}
q(\zmatrt, \thetab, \mb, \vmatr, \alphab) = q(\thetab)q(\alphab, \mb, \vmatr)q(\zmatrt).
\end{equation}
As stated in~\cite{bishop2006pattern}, $\dKL$ minimization~\eqref{eq:kl_div_joint_posterior} is equivalent to maximization of the lower bound~$\Lmc(q)$:
\begin{equation}
\label{eq:lower_bound_optimization}
\Lmc(q) = \int q(\zmatrt, \thetab, \mb, \vmatr, \alphab) \ln \left(\frac{p(\zmatrt, \zmatr,\thetab, \mb, \vmatr, \alphab)}{q(\zmatrt, \thetab, \mb, \vmatr, \alphab)} \right) \dd \thetab \dd \mb \dd \vmatr \dd \alphab \dd \zmatrt \rightarrow \max_{q\,=\,q(\thetab)q(\alphab, \mb, \vmatr)q(\zmatrt)}.
\end{equation}
To find optimal factors of~$q$ we solve~\eqref{eq:lower_bound_optimization} according to one factor of~$q$ keeping all other factors constant. It leads to the following form  of factors
\begin{equation}
\label{eq:bayes_optimal_q_predictive}
\begin{split}
\ln q(\thetab) & = \expec_{\alphab, \mb, \vmatr, \zmatrt}\bigl[\ln p(\zmatrt, \zmatr, \thetab, \mb, \vmatr, \alphab )\bigr] + \const(\thetab), \\
\ln q(\alphab, \mb, \vmatr) & = \expec_{\thetab, \zmatrt}\bigl[\ln p(\zmatrt, \zmatr, \thetab, \mb, \vmatr, \alphab) \bigr] + \const(\alphab, \mb, \vmatr), \\
\ln q(\zmatrt) & = \expec_{\alphab, \mb, \vmatr, \thetab}\bigl[\ln p(\zmatrt, \zmatr, \thetab, \mb, \vmatr, \alphab )\bigr] + \const(\zmatrt),
\end{split}
\end{equation}
where~$\const(var)$ defines some expression that does not depent on~$var$. Iterative recalculation of these factors leads to maximum because each iteration does not decrease~$\Lmc(q)$ value~\cite{bishop2006pattern}.

The likelihood~$L$~\eqref{eq:neural_classifier_log_likelihood} contains sum of exponents of random variables~$\thetab_k$ and~$\alphab$, so we can't calculate factors estimates~\eqref{eq:bayes_optimal_q_predictive} analytically. We use method of local variations~\cite{gibbs_phd} to approximate likelihood with it's upper bound. Let~$g(\xb)$ be the sum of expectations
\begin{equation}
\label{eq:sum_exp_division}
g(\xb) = \sum_{k = 1}^{K_h} \exp(x_k).
\end{equation}
Expression~$- \ln\bigl(g(\xb)\bigr)$ is a convex function (see fig. \ref{fig:upper_bound_0}), so the tangent plane through some point~$\xib$ is an upper bound for this expression
\begin{equation}
\label{eq:bayes_upper_bound_0}
y(\xb, \xib) = - \ln\bigl(g(\xib)\bigr) - \nabla \ln\bigl(g(\xib)\bigr)\T (\xb - \xib), \quad - \ln\bigl(g(\xib)\bigr) \leq y(\xb, \xib).
\end{equation}
Taking exponent from both sides of inequality~\eqref{eq:bayes_upper_bound_0} we get upper bound of one over~$g(\xb)$
\begin{equation}
\label{eq:bayes_upper_bound}
\frac{1}{g(\xb)} \leq \frac{1}{g(\xib)} \exp\left(\sum_{k = 1}^{K_h}\frac{\exp(\xi_k)}{g(\xib)} (\xi_k - x_k) \right).
\end{equation}
The index of exponent on the right side of~\eqref{eq:bayes_upper_bound} is a linear function of~$\xb$. Product of this bound and density functions from exponential family leaves it inside exponential class and makes calculation of expectation straightforward.

We get an upper bound of~$\Lmc(q)$ using constructed approximation~\eqref{eq:bayes_upper_bound} of softmax denominator for each document~$\xb_n$:
\begin{equation}
\Lmc(q) \leq \hat{\Lmc}(q, \xib),
\end{equation}
where~$\xib =  \{\xib_n\}$ is the set of variational parameters. We minimize~$\hat{\Lmc}(q, \xib)$ according to~$\xib$ to find the closest upper bound of~$\Lmc(q)$.


Optimal factors~\eqref{eq:bayes_optimal_q_predictive} calculated for joint model~\eqref{eq:bayes_prob_model_predictive} with softmax approximation~\eqref{eq:bayes_upper_bound} have the following form
\begin{equation}
\label{eq:bayes_posterior_joint_2}
\begin{split}
q(\zmatrt, \thetab, \mb, \vmatr, \alphab) & = q(\alphab)\prod_{k = 1}^{K_h}q(\thetab_k)q(\mb_k| \vmatr_k)q(\vmatr_k)\prod_{t = 1}^{|T|}q(\zt_{tk}), \\ 
q(\alphab) & \sim \Nmc(\alphab_0, a^{-1}\imatr), \\
q(\thetab_k) & \sim \Nmc(\mb_{0k}', (\nu' \vmatr_k)^{-1}), \\
q(\mb_k | \vmatr_k)q(\vmatr_k) & \sim \Nmc(\mb_{0k}, (b'\vmatr_k)^{-1})\Wmc(\wmatr_k, \nu'), \\
q(\zt_{tk}) & \sim \dBern(p_{tk}).
\end{split}
\end{equation}
Parameters~$\nu'$ and~$b'$ equals $\nu' = \nu + 1, \quad b' = 1 + b$, and parameters~$ \mb_{0k}, \wmatr_k, \alphab_0, \mb_{0k'}, p_{tk}$ are recalculated iteratively using
\begin{equation}
\label{eq:parameters_close_form}
\begin{split}
\mb_{0k} & = \frac{\expec\thetab_k + b\mb_0}{b'}, \quad \wmatr_k^{-1} = b'\mb_{0k}\mb_{0k}\T + b\mb_0\mb_0\T + \expec\bigl[\thetab_k\thetab_k\T\bigr] + \wmatr^{-1}, \\
\alphab_0 & = \frac{1}{a}\sum_{m = 1}^{|W|} \iotab_m  \sum_{k = 1}^{K_h}(\mmatr_{k}\expec \thetab_{k})_m \left(\sum_{n = 1}^N x_{nm} \hat{z}_{nk} + \sum_{t = 1}^T \xt_{tm} \hat{\hat{z}}_{tk} \right), \\
\mb_{0k}' & =  \mb_{0k} + \frac{1}{\nu'}\wmatr_k^{-1}\mmatr_k\T \expec_{\alphab}\lambdamatr \left( \sum_{n = 1}^N \xb_n \hat{z}_{nk} + \sum_{t = 1}^T\xbt_t \hat{\hat{z}}_{tk} \right), \\
p_{tk} & = \frac{\exp(\zeta_{tk})}{\exp(\zeta_{tk}) + g(\xibt_t)},
\end{split}
\end{equation}
where we defined~$\hat{z}_{nk}, \hat{\hat{z}}_{tk}$ and $\zeta_{tk}$ to make formulas uncluttered: 
\begin{equation}
\label{eq:bernouli_parameters}
\begin{split}
\hat{z}_{nk} & = \left(z_{nk} - \frac{\exp(\xi_{nk})}{g(\xib_n)}\right), \quad \hat{\hat{z}}_{tk} = \left[\expec \zt_{tk} - \frac{\exp(\xit_{tk})}{g(\xibt_t)}\left( \sum_{k' = 1}^{K_h} \expec \zt_{tk'}\right)\right], \\
\zeta_{tk} & = \xbt_t\T\expec_{\alphab}\lambdamatr \mmatr_{k}\expec\thetab_{k} + \sum_{k' = 1}^{K_h} \frac{\exp(\xit_{tk'})}{g(\xibt_t)}( \xit_{tk'} - \xbt_t\T\expec_{\alphab}\lambdamatr \mmatr_{k'}\expec\thetab_{k'}).
\end{split}
\end{equation}

\begin{minipage}[t]{0.40\linewidth}
\centering
\begin{figure}[H]
\includegraphics[width=\textwidth]{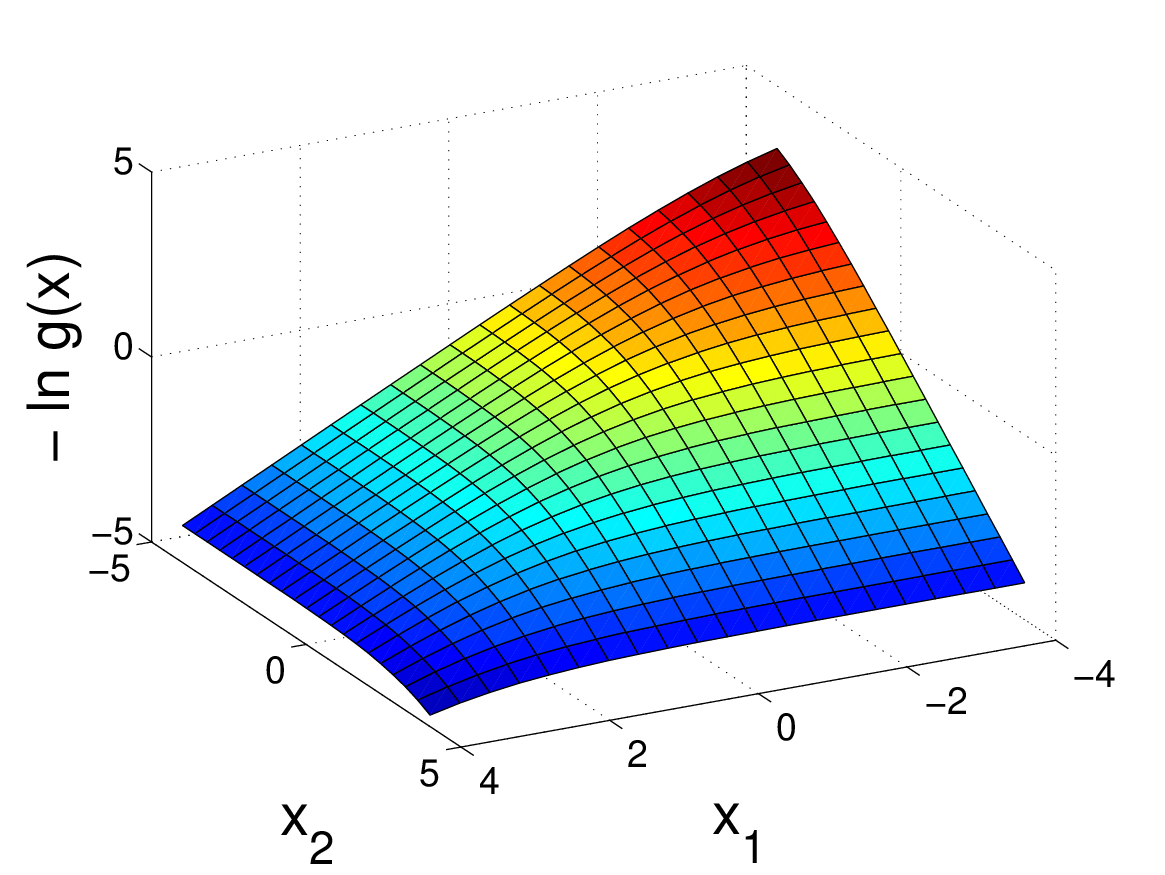}
\caption{Values of~$\tilde{g} = -\ln g(\xb)$ for two-dimensional~$\xb$.}
\label{fig:upper_bound_0}
\end{figure}
\end{minipage}
\hfill
\begin{minipage}[t]{0.45\linewidth}
\centering
\begin{figure}[H]
\includegraphics[width=\textwidth]{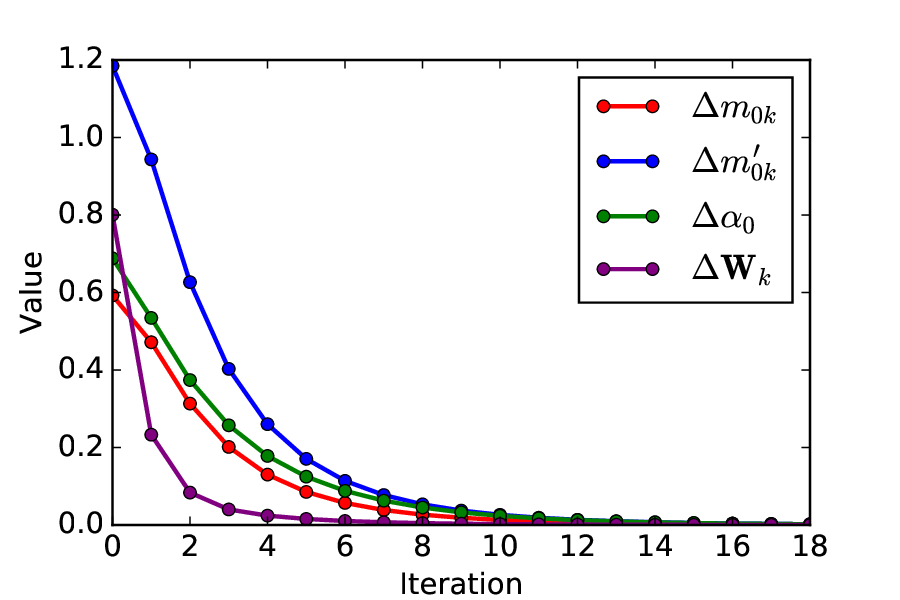}
\caption{Example of parameters convergence.}
\label{fig:parameters_convergency}
\end{figure}
\end{minipage}

\paragraph{EM algorithm for parameters optimization.} To find best approximation of the joint posterior~$p(\zmatrt, \thetab, \mb, \vmatr, \alphab| \zmatr)$ we iteratively recalculate each factor of~$q$ keeping the other fixed according to~\eqref{eq:bayes_optimal_q_predictive}. 

Likelihood upper bound~\eqref{eq:bayes_upper_bound} with variational parameters~$\xib =  \{\xib_n\}$ and~$\xibt =  \{\xibt_t\}$ allows us to calculate closed form solution~\eqref{eq:parameters_close_form} for each factor using parameters of other factors. This leads to~EM algorithm which alternate~E step to calculate~$q$ parameters using~\eqref{eq:parameters_close_form} with~M step to optimize variational parameters~$\xib, \xibt$ of~$\Lmc(q)$ upper bound.
\begin{enumerate}
\item Initialize parameters
$$
\wmatr, \nu, \mb_0, a, b, \wmatr_k = \wmatr, \nu' = \nu + 1, b' = b + 1,  \mb_{0k} = \mb_0, \xib_n.
$$
\item Calculate~$\expec \thetab_k, \expec [\thetab_k\thetab_k\T]$ according to~$q(\thetab_k)$ distributions
\begin{equation}
\begin{split}
\expec \thetab_k & = \mb_{0k}', \\
\expec [\thetab_k\thetab_k\T] & = (\nu' \wmatr_k)^{-1} +  \mb_{0k}'(\mb_{0k}')\T,
\end{split}
\end{equation}
and recalculate parameters of~$q(\mb), q(\vmatr), q(\alphab)$ factors using~\eqref{eq:parameters_close_form}. 
\item Calculate~$\expec_{\alphab} \lambdamatr$ using~$q(\alphab)$ distribution
\begin{equation}
\expec_{\alphab} \lambdamatr = \tilde{\lambdamatr} = \diag(\{\lambda_m'\}),\quad \lambda_m' = 1 + \alphab_0\T \iotab_m,
\end{equation}
and recalculate parameters of~$q(\thetab_k)$ factors using~\eqref{eq:parameters_close_form}. 
\item Optimize variational parameters~
\begin{equation}
\xi_{nk} = \xb_n\T \tilde{\lambdamatr} \mmatr_k \mb_{0k}', \quad \xit_{tk} = \xbt_t\T \tilde{\lambdamatr}\mmatr_k\mb_{0k}'.
\end{equation}
If some of parameters have changed significantly on the steps 2-4, go back to step 2.
\end{enumerate}

\paragraph{Probability of a class given document.} The optimal joint posterior approximation has the form~\eqref{eq:bayes_posterior_joint_2}, where the distribution of the class~$c_{h, k}$ label~$\zt_{tk}$ for a document~$\xbt_t$ is a bernoulli distribution with parameter~$p_{tk}$~\eqref{eq:bernouli_parameters}. The integral from joint posterior~\eqref{eq:evidence_estimation} gives bayesian estimate of cluster probability. Substitution of~\eqref{eq:bayes_posterior_joint_2} into~\eqref{eq:evidence_estimation} gives straightforward estimate of probability~$p(\zt_{tk} = 1|\xbt_t) = p_{tk}$. For each document relevance operator R ranks clusters according to this estimate.

\section{Computational Experiment}
To test the proposed approach and compare it with well-known methods we solve a hierarchical classification task for two text collections: abstracts of EURO conference and web-sites of industry companies.

\paragraph{Collection of EURO abstracts.} We used programs of scientific conference~EURO from~2006 till~2016~\cite{EURO_abstracts}. To unify data from conferences of different years and to build a single structure for collection we used the following procedure.
\begin{enumerate}
\item[1.] Take an expert cluster structure of EURO~2016 as a Base~(fig.~\ref{fig:euro_structure}).
\item[2.] For each cluster~$c$ of EURO~2010-2015 conferences search for the same cluster in the Base structure. If the one is found, merge~$c$ it with, if there is no such cluster, added~$c$ as a new one to the Base structure.
\item[3.] For each cluster~$c$ of EURO 2006-2009 conferences search for the same cluster in the Base structure. If the one is found, merge~$c$ it with, if there is no such cluster, discard all documents from~$c$.
\end{enumerate}
The joint collection contains~$|D| = 15527$ documents, dictionary contains~$|W| = 24304$ words, the Base hierarchical structure consists of~$K_2 = 26$ clusters of the second level (Area) and~$K_3 = 264$ clusters of the third level (Stream).

\paragraph{Ranking results for unlabeled documents.} For ranking experiment we considered Area and Stream levels of hierarchy. We constructed relevance operator~$R$ using proposed weighted hierarchical similarity function hSim~\eqref{effective_object_to_cluster_similarity} and used EM algorithm from the section~\ref{sec:bayes} to optimize it's parameters on the traning subset. Results of this function were compared with other algorithms of hierarchical ranking: 1)~hierarchical naive bayes hNB~\cite{DBLP:conf/icml/McCallumRMN98}, 2)~probabilistic regularized model SuhiPLSA~\cite{DBLP:conf/iconip/KuznetsovCAGS15} and 3)~hierarchical multiclass svm~\cite{hierarchicalSVM2007}.

We divided collection~$D$ into two parts: train~$D_{\Vmc}$ and test~$D_{\Tmc}$ in different proportions: size of the training subset~$|D_{\Vmc}|$ varied from~$500$ documents to~$10000$. Size of the testing subset~$D_{\Tmc}$ was fixed,~$|D_{\Tmc}| = 5000$. Each algorithm was trained on the~$D_{\Vmc}$ and returned a ranked list of clusters for a given document. Quality of the algorithms was measured using area under cumulative histogram~$\AUCH$~\eqref{eq:auch}.

\begin{figure}[h]
\begin{center}
\includegraphics[width=0.7\textwidth]{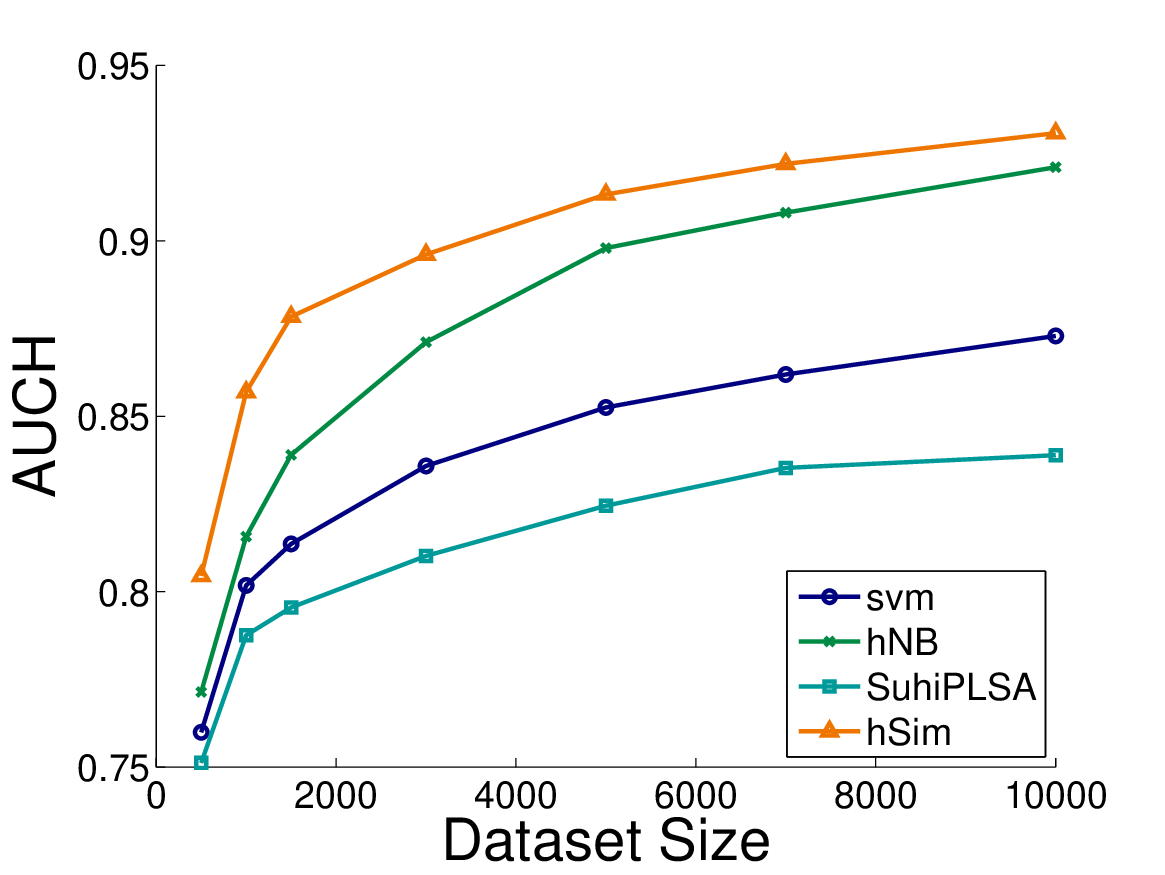}
\caption{Test AUCH quality dependence from training sample size for svm, hNB, suhiPLSA, hSim algorithms.}
\label{fig:algo_comp_area_stream_dataset_size}
\end{center}
\end{figure}

\begin{figure}[H]
\begin{minipage}[t]{0.49\linewidth}
\centering
\includegraphics[width=\textwidth]{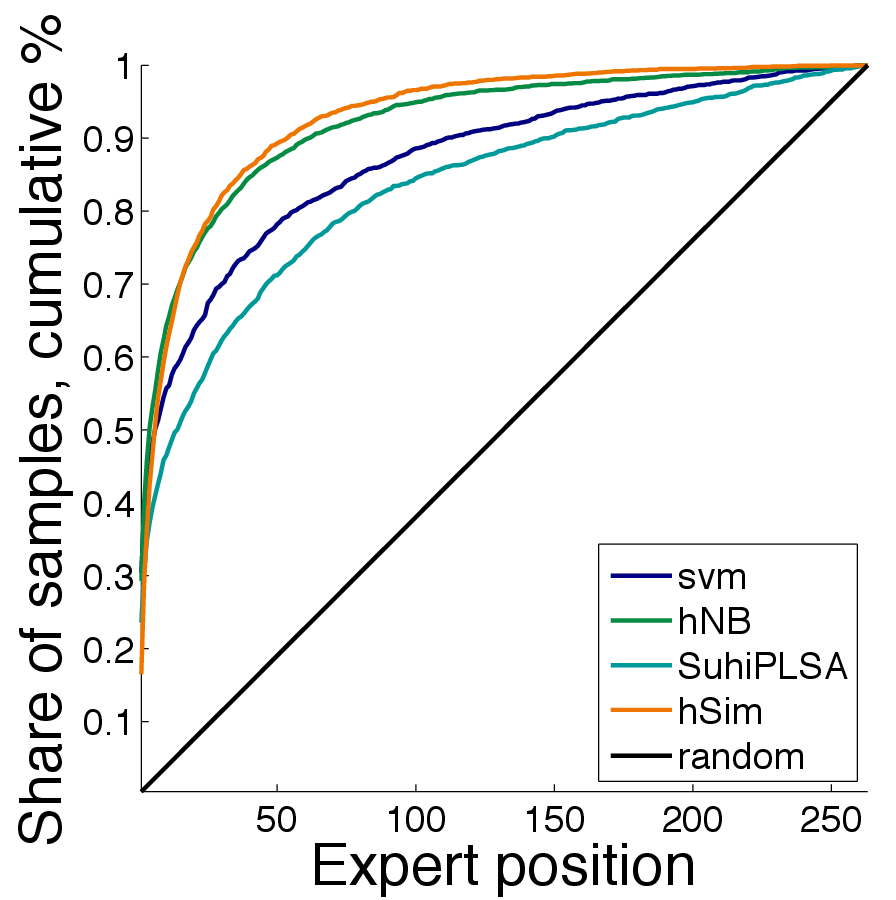}\\
{Collection of EURO conference abstracts,~$10000$ training objects.}
\end{minipage}
\begin{minipage}[t]{0.49\linewidth}
\centering
\includegraphics[width=\textwidth]{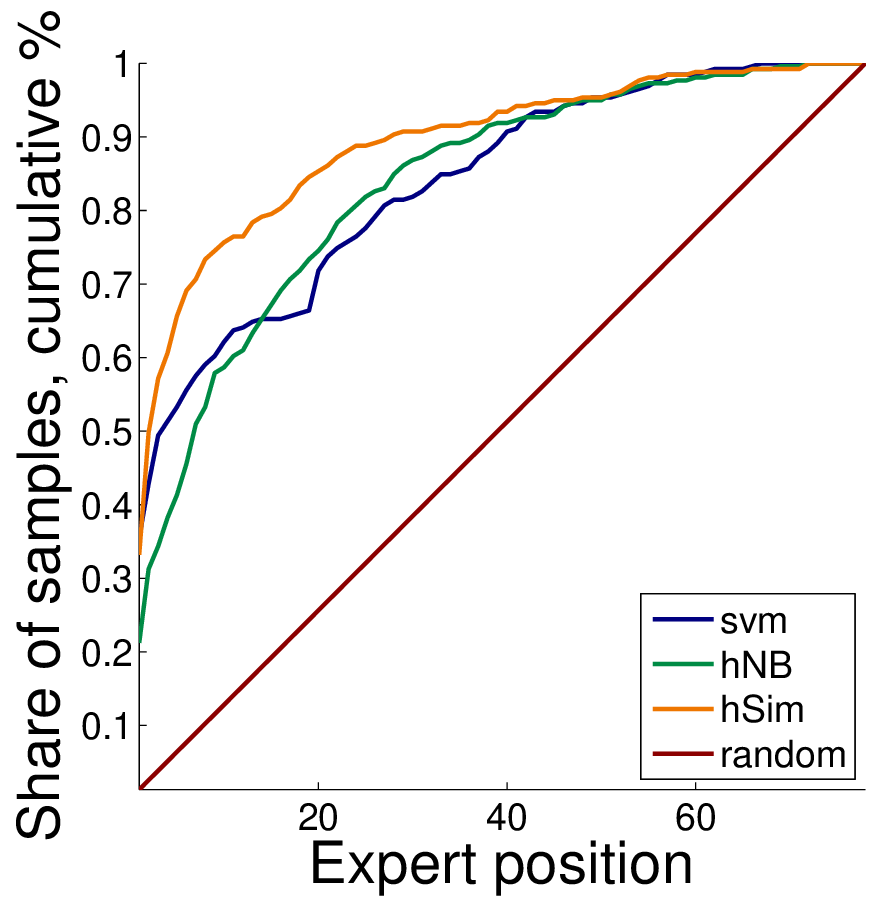}\\
{Collection of industry companies web sites.}
\end{minipage}
\captionof{figure}{Envelope curves of cumulative histograms for different algorithms.}
\label{fig:algo_comp_area_stream_industry}
\end{figure}

The fig.~\ref{fig:parameters_convergency} shows the convergence of parameters during optimization with~EM algorithm from section~\ref{sec:bayes}. The table~\ref{tab:algo_comp_stream} contains values of~AUCH for all algorithms and sizes of training sample. Bold values corresponds to the best statistically equivalent values for each training sample size. Fig.~\ref{fig:algo_comp_area_stream_dataset_size} shows the table data in the charts format. Proposed hSim algorithm showed the best results. Fig.~\ref{fig:algo_comp_area_stream_industry}a. shows the envelope curve for cumulative histogram~\eqref{eq:histogram} for training sample size~$|D_{\Vmc}| = 10000$.

\begin{table}[!h]
\captionof{table}{Ranking quality AUCH~\eqref{eq:auch} of the different algorithms and training set sizes.}	
\label{tab:algo_comp_stream}
\centering
\begin{tabular}{|l|c|c|c|c|c|c|c|}
\hline
{Algorithm}{Training  set size $|D_{\Vmc}|$} & $500$ & $1000$ & $1500$ & $3000$ & $5000$ & $7000$ & $10000$ \\
\hline
svm & $0.76$  &  $0.80$  &  $0.81$  &  $0.84$ &   $0.85$ &   $0.86$  &  $0.87$\\
\hline
hNB & $0.77$  &  $0.82$ &   $0.84$ &   $0.87$  &  $0.90$  &  $0.91$   & $0.92$\\
\hline
suhiPLSA & $0.75$  &  $0.79$ &   $0.80$ &   $0.81$  &  $0.82$  &  $0.84$   & $0.84$\\
\hline
hSim & $0.80$  &  $\mathbf{0.86}$  &  $\mathbf{0.88}$  &  $\mathbf{0.90}$  &  $\mathbf{0.91}$ &   $\mathbf{0.92}$  &  $\mathbf{0.93}$\\
\hline
hSimWV & $\mathbf{0.82}$  &  $\mathbf{0.86}$  &  $0.87$ &   $\mathbf{0.89}$  &  $\mathbf{0.92}$ &   $\mathbf{0.92}$ &   $0.92$\\
\hline
\end{tabular}
\end{table}

\begin{figure}[!h]
\begin{minipage}[h]{0.49\linewidth}
\center{\includegraphics[width=0.95\textwidth]{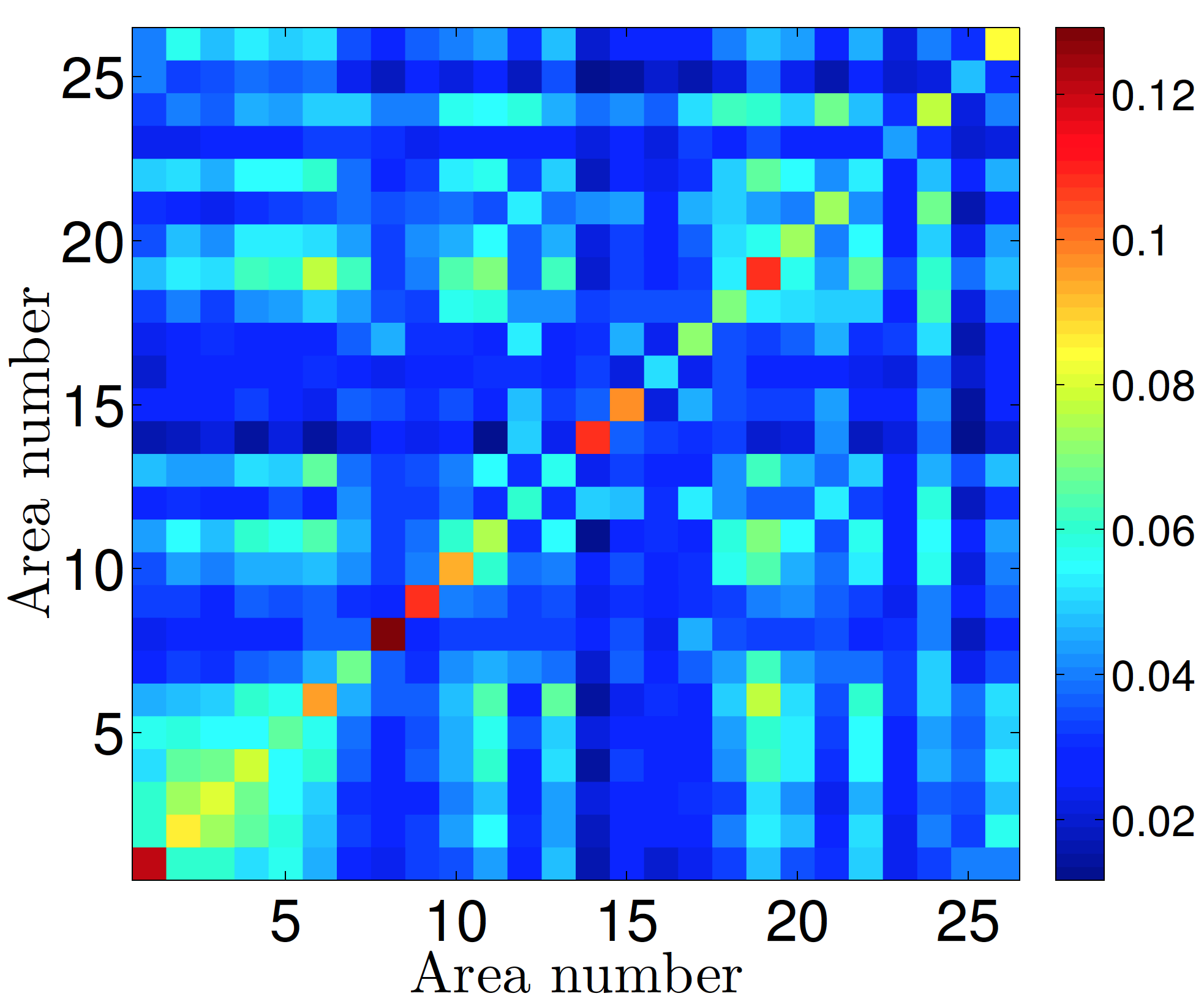}\\
Same importance for all words, $\lambdamatr = \imatr$.}
\end{minipage}
\hfill
\begin{minipage}[h]{0.49\linewidth}
\center{\includegraphics[width=0.95\textwidth]{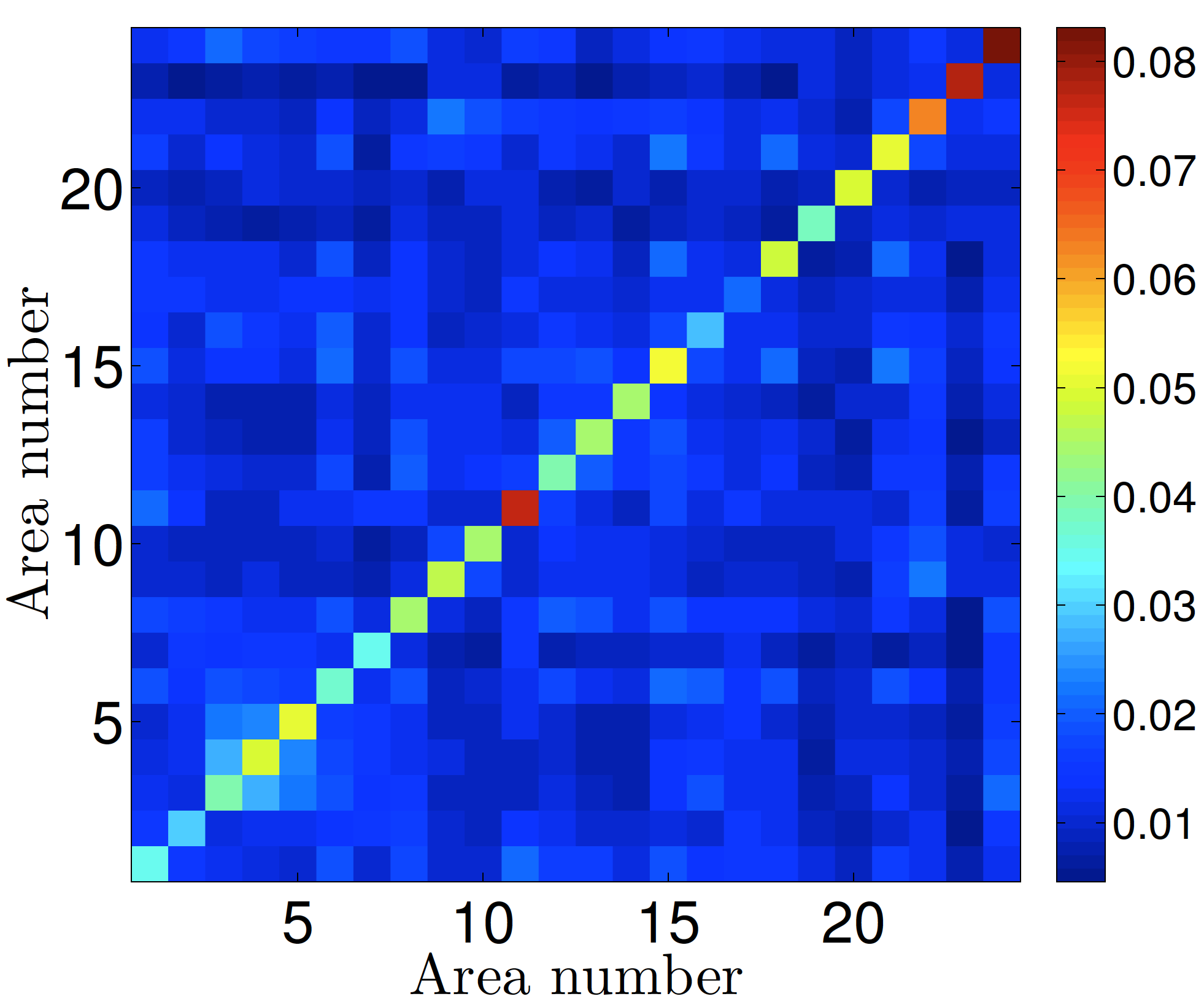}\\
Optimized importance of words, $\lambdamatr = \lambdamatr^*$.}
\end{minipage}
\captionof{figure}{Pairwise similarities of Area level clusters with entropy model (right chart) and without it (left chart).}
\label{fig:lambda_comparison_new}
\end{figure}

Fig.~\ref{fig:lambda_comparison_new} demonstrates the effect of entropy model. It visualize matrix of pair similarity of expert clusters on the Area level of the hierarchy. We suppose that expert clustering is an optimal solution, so similarity function should separate intracluster similarity and intercluster similarity well. Right part of the fig.~\ref{fig:lambda_comparison_new} shows the values of weighted cluster similarity that uses~$\lambdamatr = \lambdamatr^*$ with optimal entropy model parameter~$\alphab$, the left part of the fig.~\ref{fig:lambda_comparison_new} shows the values of clusters similarity without optimization,~$\lambdamatr = \imatr$. We can see from the figures that intracluster similarities (diagonal elements of the matrix) became greater than intercluster similarities (non-diagonal elements) after optimization. Optimal~$\lambdamatr^*$ corresponds to~$0.047$ average intracluster similarity and~$0.012$ average intercluster similarity.

\paragraph{Collection of industrial companies web sites.} In this collection each web site is represented by set of HTML pages. We merge all pages into one and remove all special symbols and tags to form a single text document for each site. The final collection contains~$|D| = 1036$ documents, dictionary contains~$|W| = 18775$ words, hierarchical structure contains~$K_2 = 11$ clusters of the second level and~$K_3 = 78$ clusters of the third level.

Training subset~$D_{\Vmc}$ consists from~$750$ documents, and test subset~$D_{\Tmc}$ consists from remaning~$286$ documents. We compare results of the proposed weighted hierarchical similarity function hSim with~1)~hierarchical naive bayes hNB~\cite{DBLP:conf/icml/McCallumRMN98} and~2)~hierarchical multiclass svm~\cite{hierarchicalSVM2007}. Table~\ref{tab:algo_comp_stream_industry} shows the~AUCH~\eqref{eq:auch} quality criterion values for these algorithms. Fig.~\ref{fig:algo_comp_area_stream_industry}b. shows corresponding envelope of the cummulative histograms~\eqref{eq:histogram}.
 
\begin{table}[H]
\captionof{table}{$\AUCH$~\eqref{eq:auch} values for different algorithms. Collection of industry companies web sites.}	
\label{tab:algo_comp_stream_industry}
\centering
\begin{tabular}{|l|c|}
\hline
Algorithm & AUCH \\
\hline
svm & $0.83$ \\
\hline
hNB  & $0.83$ \\
\hline
hSim & $\mathbf{0.89}$ \\
\hline
\end{tabular}
\end{table}
 
\section*{Conclusion}
In this paper we solve hierarchical text classification task for partly-labeled collections with tree cluster structure given by the experts. To find relevance of the clusters to the given document we propose weighted hierarchical similarity function of a document and a branch of the cluster structure. It allows to rank the whole branches of the hierarchy instead of using common top-down approach. Proposed function contains two sets of parameters: words importance for classification and weight vectors for each branch of the cluster tree. To estimate importance of the words we propose model that calculates word's importance using it's entropy. 

To use effective optimization techniques we propose joint probabilistic model of document classes, parameters and hyperparameters. Variational bayesian inference and likelihood upper bound allows us to approximate the joint posterior distribution of unlabeled document classes and parameters, and calculate bayesian estimate of a class probability given document. The proposed relevance operator ranks clusters according to probability estimate. We compare results of our approach with hierarchical multiclass SVM, hierarchical naive bayes and PLSA with ARTM regularization on two types of text collections: abstracts of the major conference EURO and web sites of industry companies. Proposed approach showed comparable results on both collections.

For the future work we are going to use other types of likelihood bounds like quadratic lower bound; compare the variational inference results with sampling techniques; generalize hierarchical similarity to other types of cluster structures like directed acyclic graphs. 

\bibliographystyle{unsrt}
\bibliography{hierarchical_similarity}
\end{document}